\documentclass[sn-mathphys, UTF-8]{sn-jnl}

\jyear{2021}%
\usepackage{array}
\usepackage{booktabs}
\usepackage[justification=centering]{caption}
\usepackage{amsmath}
\usepackage{ulem}
\usepackage{cleveref}
\theoremstyle{thmstyleone}%
\theoremstyle{thmstyletwo}%

\theoremstyle{thmstylethree}%

\begin{document}

\title[]{SIFTER: A Task-specific Alignment Strategy for Enhancing Sentence Embeddings}

\author[1]{\fnm{Chao} \sur{Yu}}\email{yuchao@shu.edu.cn}

\author*[2]{\fnm{Wenhao} \sur{Zhu}}\email{whzhu@shu.edu.cn}

\author[1]{\fnm{Chaoming} \sur{Liu}}

\author[1]{\fnm{Xiaoyu} \sur{Zhang}}

\author[1]{\fnm{Qiuhong} \sur{Zhai}}


\affil[1]{\orgdiv{School of Computer Engineering and Science}, \orgname{Shanghai University},  \street{No 99, Shangda Road}, \city{Shanghai, 200444}, \country{China}}

\affil[2]{ \orgname{Information technology office}, \orgname{Shanghai University}, \street{No 99, Shangda Road}, \city{Shanghai, 200444}, \country{China}}

\abstract{The paradigm of pre-training followed by fine-tuning on downstream tasks has become the mainstream method in natural language processing tasks. Although pre-trained models have the advantage of generalization, their performance may still vary significantly across different domain tasks. This is because the data distribution in different domains varies. For example, the different parts of the sentence `He married Smt. Dipali Ghosh in 1947 and led a very happy married life' may have different impact for downstream tasks. For similarity calculations, words such as `led' and `life' are more important. On the other hand, for sentiment analysis, the word `happy' is crucial. This indicates that different downstream tasks have different levels of sensitivity to sentence components. Our starting point is to \underline{$\mathbf{s}$}cale \underline{$\mathbf{i}$}n\underline{$\mathbf{f}$}ormation of the model and data according to the specifics of downstream \underline{$\mathbf{t}$}asks, \underline{$\mathbf{e}$}nhancing domain information of \underline{$\mathbf{r}$}elevant parts for these tasks and reducing irrelevant elements for different domain tasks, called $\mathbf{SIFTER}$. In the experimental part, we use the SIFTER to improve SimCSE by constructing positive sample pairs based on enhancing the sentence stem and reducing the unimportant components in the sentence, and maximize the similarity between three sentences. Similarly, SIFTER can improve the gate mechanism of the LSTM model by short-circuiting the input gate of important words so that the LSTM model remembers the important parts of the sentence. Our experiments demonstrate that SIFTER outperforms the SimCSE and LSTM baselines.}




\maketitle

\section{Introduction}\label{sec1}

Domain shift in Natural Language Processing is a common issue, referring to the phenomenon where a model trained in one domain performs poorly when applied to a different domain. There exists a solution via pretraining-finetuning, which enables the model to adapt to a new environment by utilizing a small amount of downstream data. This approach has become the mainstream method following the advent and subsequent evolution of models such as ELMo, BERT, and GPT. However, a problem with this method is that the pretrained model and the downstream task data do not sufficiently reflect the characteristics of the domain-specific task.

As shown in Table 1, SpeBert (Liu et al., 2022) demonstrate that sentence main parts (subject, predicate, and object) is important for semantic similarity tasks. Meanwhile, the sentence other parts (adverbial and complement) are more important than the main parts in the sentiment analysis task. AugCSE (Tang et al., 2022) has demonstrated that it can improve the performance of natural language inference tasks, while other transfer tasks lead to performance degradation. Therefore, it is not easy to find a method that is suitable for a variety of downstream tasks. We need to make corresponding changes for different downstream tasks.

Therefore, we propose an idea, which involves increasing and decreasing the information of the model and data to adapt to the downstream tasks, called SIFTER. In tasks such as semantic similarity and sentiment analysis, we adaptively scale the information of the data and model for downstream tasks, enabling the model to acquire more domain-specific signals during the learning process.

\begin{figure}[!htb]
	\centering
	\includegraphics[width=0.8\textwidth]{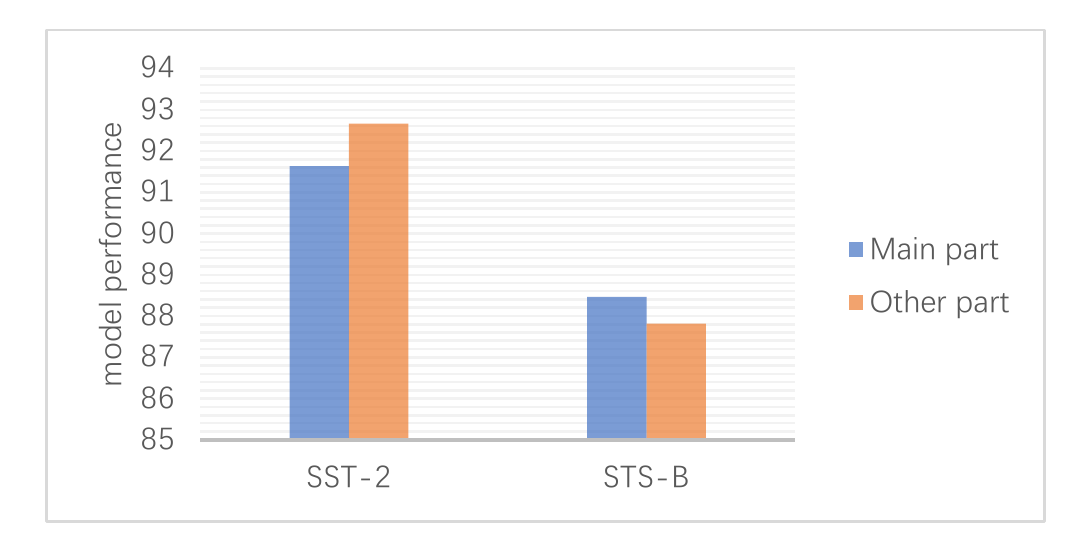}
	\caption{Impact of enhanced sentence parts on downstream tasks.}\label{<fig1>}
\end{figure}

Adopting sentence representations is a common approach on the semantic similarity task, how to express sentence vectors is still a problem. The reason is that sentence vectors learned through pretrained language models exhibit anisotropy and word frequency biases, which makes the computation using cosine similarity very poor. Contrastive learning can alleviate these phenomenon. Some recent works use contrastive objective to fine tune pre-trained language models without any labeled data (Wu et al., 2020; Giorgi et al., 2020; Gao et al., 2021; Wu et al., 2021). SimCSE (Gao et al., 2021) proposes a simple and efficient way of data augmentation, using dropout mask as the minimum positive sample pair construction. ESimCSE (Wu et al., 2021a) proposes random word repetition to construct pairs of positive samples based on the SimCSE method, and maintains a queue of negative samples using a momentum encoder.

For semantic similarity tasks, the main part information of a sentence is especially important, while elements like conjunctions and articles are not essential in these tasks. Therefore, following the strategy of SIFTER, we constructed two positive samples suitable for SimCSE on semantic similarity tasks. One sample involves a sentence with enhanced main information, and the other involves a sentence with unnecessary information removed, and negative instances are obtained from other sentences in the batch.

To validate the efficacy of SIFTER in adapting models for downstream tasks, we have opted for the Long Short-Term Memory (LSTM) model to substantiate our theory. In sentiment analysis tasks, sentiment words typically directly denote the emotional tendency of the sentence, and play a pivotal role in the predictive accuracy of the model. The gate mechanism of LSTM can decide which information to discard and which to retain. Hence, by tweaking the gate mechanism of LSTM, we can channel the information of sentiment words directly into the LSTM's memory cell. This approach can effectively preserve the information of sentiment words, encouraging the model to pay greater attention to these words during sentiment analysis tasks, thereby enhancing the predictive accuracy of the model.

In the experiments, we demonstrate that validity of SIFTER on seven standard semantic textual similarity (STS) tasks \cite{2012semeval, 2013sem, 2014semeval, 2015semeval, 2016semeval, 2017semeval, 2014sick} and Sentence-level sentiment analysis (SA) tasks \cite{2013recursive}. Experimental results show that the method proposed in this paper improves the sentence representation and sentiment  classification accuracy. 

Contributions. (1) We proposed the SIFTER strategy, which means that the specific operations should be closely aligned with the characteristics of downstream tasks, enhancing necessary information in the data and models, while eliminating unnecessary information. (2) We demonstrate that the effectiveness of SIFTER idea in constructing positive samples on semantic similarity tasks and achieve beyond baseline results. (3) We also show that SIFTER can apply modifications to the LSTM gate mechanism on sentiment analysis tasks.

\section{Related Work}\label{sec2}

\subsection{Transfer learning}\label{sec2.1}

Domain shift is a common phenomenon in the fields of Natural Language Processing. It describes a scenario where the performance of a model may be impacted if the distributions of the training and testing data are different. This phenomenon often occurs when a model transitions from one domain to another, such as shifting from semantic similarity to sentiment analysis. Given the potential for significant differences in data distribution and features between domains, this could lead to a decline in the model's performance in the new domain. Transfer Learning is one such method to address the problem of domain shift. Pretrained language models like BERT, GPT, and others serve as typical examples of Transfer Learning. They are pre-trained on large volumes of text data, then fine-tuned for specific tasks.

\subsection{Sentence Embedding}\label{sec2.2}

A large body of research on word representations has made progress \cite{2020enhanced, 2021learning}. Most applications require further extensions of word representations to sentence representations for downstream tasks in NLP. Prior to the advent of BERT, many methods to obtain sentence embeddings were obtained through RNN-based models. 

For example, some previously developed approaches include skip thought \cite{2015skip} and quick thought \cite{2018efficient}, which use unlabeled datasets for unsupervised sentence representation learning. Sentence representation methods learning can also be performed by using publicly annotated datasets (e.g., NLI \cite{2015large, 2017broad}) for supervised training, such as inferSent \cite{2017supervised}. After the advent of BERT, researchers started to obtain sentence representations based on the BERT model. Sentence-BERT \cite{2019sentence} uses the siamese structure to output semantically valuable sentence embeddings on labeled datasets. However, the above methods have shortcomings, and unsupervised methods often do not work well. Supervised training requires a large number of labeled sentence pairs, which are often unavailable in many real-world scenarios.

\subsection{Contrastive Learning}\label{sec2.3}

Contrastive learning has achieved great success in the field of computer vision, and the most famous approaches of which are MoCo \cite{2020momentum} and SimCLR \cite{SimCLR}. These methods use different data enhancement methods (such as rotation, cropping, cutout, etc.) to identify the similar objects and distinguish different images. The essence of such a method is to make the high-dimensional embedding representations of pairs of similar samples close to each other, while pairs of dissimilar samples are kept far apart. In the NLP field, contrastive learning has been applied to learn sentence embeddings \cite{Clear, Declutr, Simcse, ESimcse, 2021semantic}. 

Among the related approaches, the simplest work is SimCSE. SimCSE follows the framework of SimCLR and uses the dropout mask on the given sentence as the minimal data augmentation and on the other sentences in the batch as negative samples. Furthermore, InfoNCE (NT-Xent) is used as the training loss. ESimCSE demonstrates that sentences with the same length are likely to be considered similar sentences by the SimCSE model, because of the influence of positional encoding in transformers \cite{2017attention}. To remove the bias of the SimCSE model, ESimCSE was proposed with a word repetition strategy and a momentum encoder using MoCo, and ESimCSE outperforms SimCSE on semantic similarity tasks.

\section{Improving the SimCSE Via SIFTER}\label{sec3}

In this section, we will first provide a brief introduction to the concept of contrastive learning, followed by a detailed explanation of the SimCSE method. Next, we will apply the STRIVE strategy to construct positive samples for improving SimCSE, specifically for semantic similarity tasks.

\subsection{Contrastive Learning and Unsupervised SimCSE}\label{sec3.1}

Contrastive learning is a self-supervised way to learn effective representations. It works by narrowing the distances between similar examples and pushing the distances between dissimilar examples farther away through comparisons. The key to contrastive learning is how to construct positive and negative pairs. Thus, we introduce SimCSE. SimCSE follows the SimCLR framework. Given a set of sentences $\left \{ x_{i}  \right \} _{i=1}^{M}$, SimCSE creates positive examples $ \left \{ x_{i}^{+}  \right \} _{i=1}^{M} $ for the sentence set by applying different dropout masks. Let $ f_{\theta } \left ( \cdot  \right ) $ be the pretrained language model encoder. Next, we can obtain the sentence embeddings $ h_{i} = f_{\theta }\left (  x_{i} \right ) $ and $ h_{i}^{+}  = f_{\theta }\left (  x_{i}^{+} \right ) $ for $ x_{i} $ and $ x_{i}^{+} $. The SimCSE training objective is:

\begin{equation}
\ell_i = - \log \frac{e^{{sim}({h}_i, {h}_i^{+}) / \tau }}{\sum_{j = 1}^Ne^{{sim}({h}_i, {h}_j^{+}) /\tau}} \label{eq1}
\end{equation}

The purpose of formula (1) is to enhance the similarity between $ x_{i} $ and $ x_{i}^{+} $, where N is the size of the minibatch, $\tau$ is a temperature parameter and $ sim\left ( u,v \right ) $ denotes the dot product between the L2 normalized versions of u and v.

\begin{figure}
	\centering
	\includegraphics[width=1\textwidth]{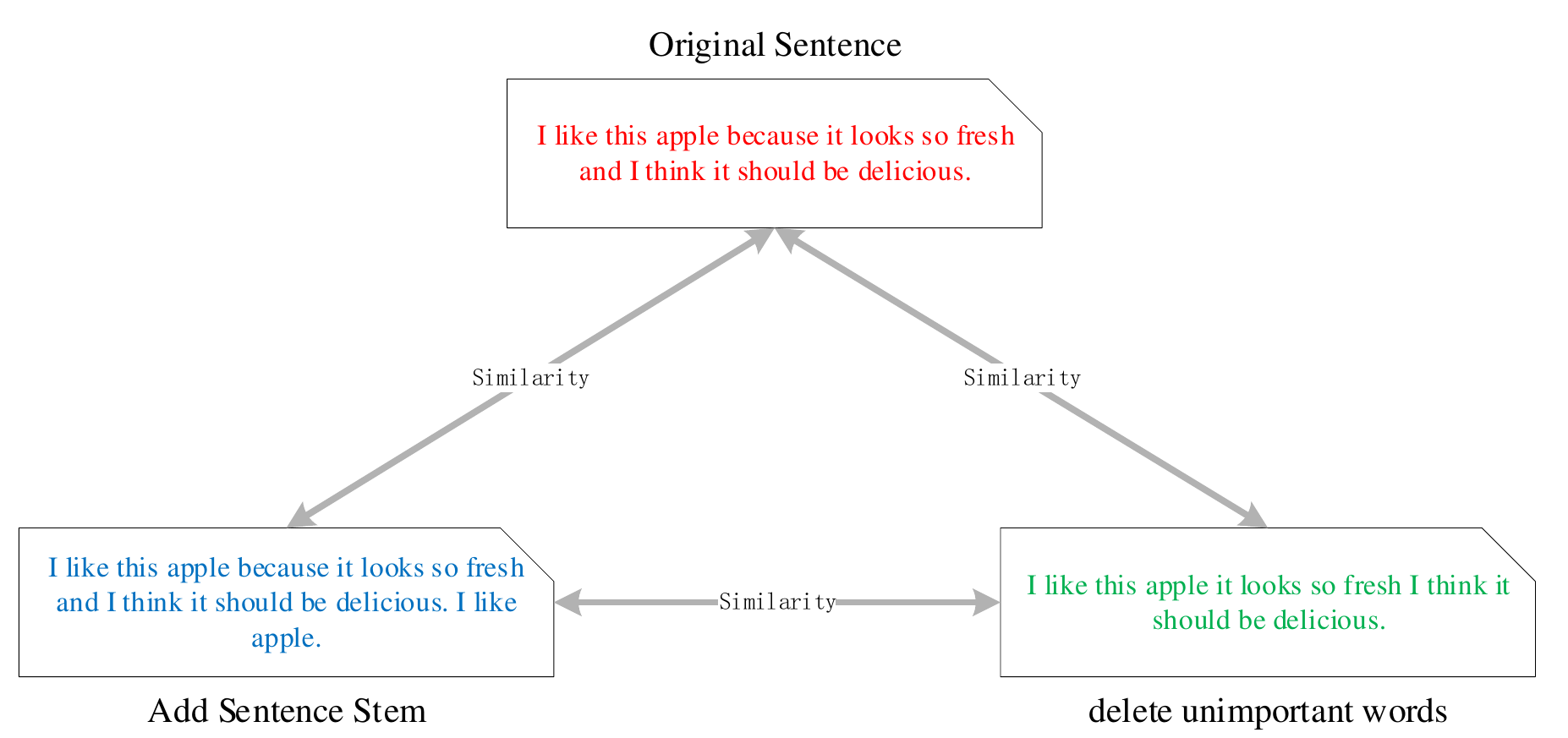}
	\caption{Reinforcing the similarities between the original sentence and the sentence with the stem added and the sentence with the useless information removed.}\label{<fig1>}
\end{figure}

\subsection{Data Augmentation for Semantic Similarity Tasks}\label{sec3.2}

SimCSE and ESimCSE studies find that random deletion of words in sentences leads to degradation of model performance. Meanwhile, SpeBert study show that the sentence backbone information is beneficial for semantic similarity tasks. So, we can enhance the main information of a sentence as a positive sample for SimCSE, while constructing another positive sample by reducing semantically irrelevant words, such as articles and conjunctions. The operation is simple: we append the backbone information of the original sentence after the original sentence. The second is to remove conjunctions and pronouns from articles and parts of sentences. The following is a concrete implementation.

(1) Add sentence backbone: For short sentences, we use StanfordCoreNLP tools to parse the sentence, obtain a dependency syntax tree of the sentence, and use the first two layers of the syntax tree as the sentence backbone. Since it is difficult to find the sentence backbone, so we use Stanford OpenIE to extract relation triples from sentences, randomly select a triple as the backbone of a sentence and add it to the back of the original sentence.

(2) Delete useless words: In general, articles do not change the semantics of sentences, while conjunctions contain less semantic information. We remove these words to achieve minimal semantic change.

Let us take the example given by ESimCSE for demonstration purposes, \cref{tab2} shows that the first data augmentation method is to add ``I like apple.'' as the sentence backbone after the original sentence, the second method is to add the conjunctions ``because'' and ``and'' and remove them from the sentence. Both results retain the meaning of the original sentence fairly well.

\renewcommand\arraystretch{1.5}
\begin{table}[h]
	\caption{Examples of constructing positive sentences. We use the unsupervised SimCSE model to compute the similarity scores between the given sentences.}\label{tab2}%
	\begin{tabular}{|l|l|c|}
		
		\hline
		Method            & Text                                                                                                                       & similarity \\ \hline
		original sentence & \begin{tabular}[c]{@{}l@{}}I like this apple because it looks so fresh and  I \\ think it should be delicious.\end{tabular} & 1.0        \\ \hline
		word repetition &
		\begin{tabular}[c]{@{}l@{}}I \textbf{I} like this apple \textbf{apple} because it looks \textbf{looks} \\ so fresh \textbf{fresh} and I think it should be delicious \\ \textbf{delicious}.\end{tabular} &
		0.987 \\ \hline
		random deletion   & \begin{tabular}[c]{@{}l@{}}I \sout{like} this apple because it looks so \sout{fresh} and \\ I think it should be \sout{delicious}.\end{tabular} & 0.792      \\ \hline
		\textbf{add sentence backbone} &
		\begin{tabular}[c]{@{}l@{}}I like this apple because it looks so fresh and \\ I think it should be delicious. \textbf{I like apple.}\end{tabular} &
		0.988 \\ \hline
		\textbf{delete useless words} &
		\begin{tabular}[c]{@{}l@{}}I like this apple \sout{because} it looks so fresh \sout{and} \\ I think it should be delicious.\end{tabular} &
		0.983 \\ \hline
	\end{tabular}
\end{table}

Intuitively, the original sentence X should be similar to sentence Y with a sentence stem added and sentence Z with useless words removed. As shown in \cref{<fig1>}, we enforce the semantic similarities between X, Y and Z during model training.

\subsection{Model Architecture}\label{sec3.3}

\begin{figure}
	\centering
	\includegraphics[width=1\textwidth,height=0.6\textwidth]{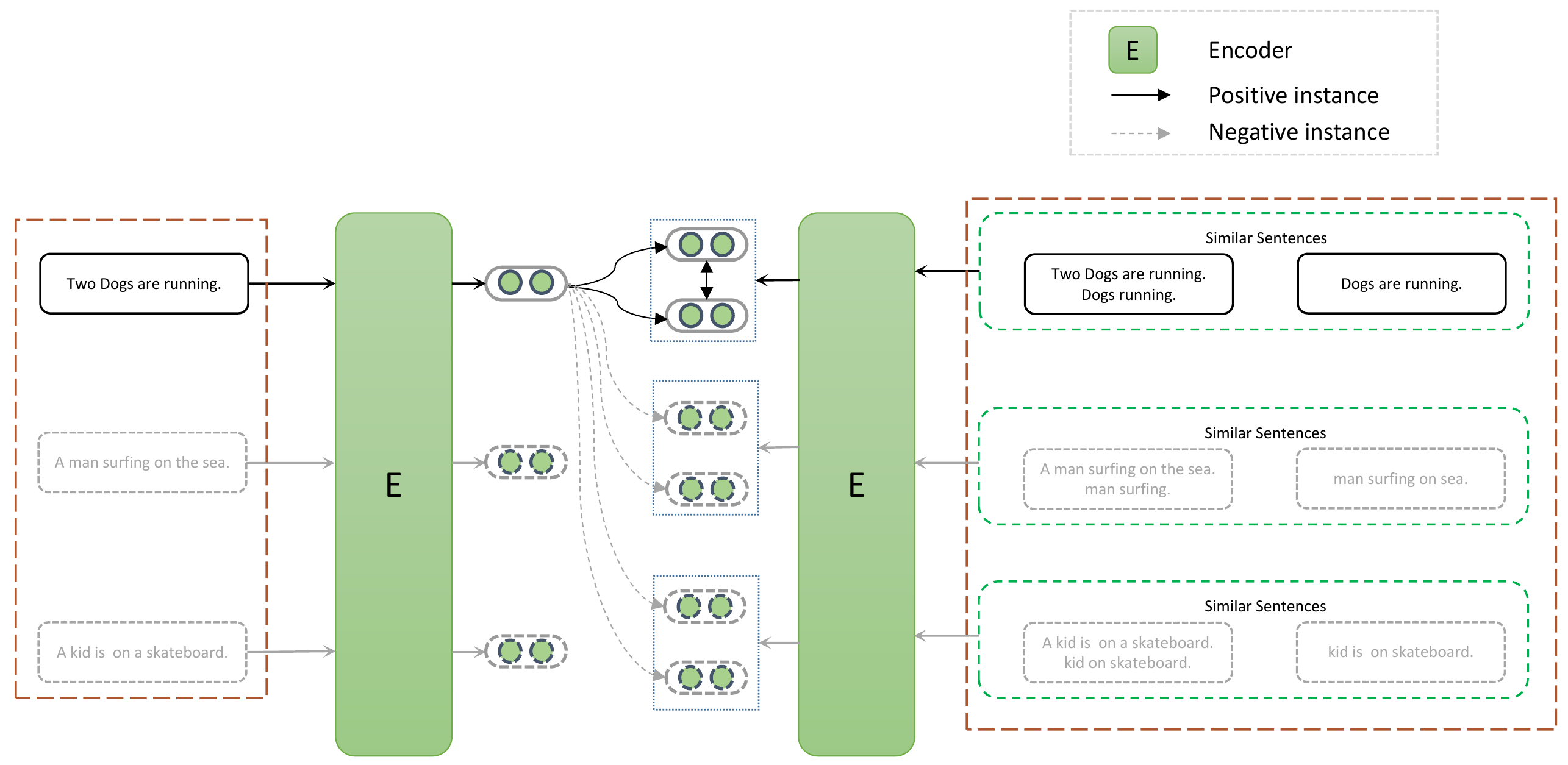}
	\caption{SIFTER-SimCSE method diagram. In contrast to the unsupervised SimCSE approach, we construct pairs of positive samples in two ways. The left-hand side is the sentence representation obtained from the original sentence by a pretrained language encoder, and the right-hand side is the sentence representation obtained from the corresponding two positive samples by the same encoder. The three representations are close to each other on the hypersphere but far from the examples of positive samples constructed from other sentences in the minibatch.}\label{<fig2>}
\end{figure}

Our model framework is the same as that of the unsupervised SimCSE method. The difference is that the proposed model is trained with two data augmentation methods for the semantic similarity task. During training, we explicitly let the sentence representation learn which components contribute to the meaning of the sentence and which components are noisy. The overview of our framework is shown in \cref{<fig2>}.

Given a sentence set $ \left \{ x_{i}  \right \} _{i=1}^{M} $, we can build a sentence set $ \left \{ x_{i}^{Y+}  \right \} _{i=1}^{M} $ after adding the backbone and a sentence set $ \left \{ x_{i}^{Z+}  \right \} _{i=1}^{M} $ after removing useless information, where M is the size of the sentence set. We can obtain the hidden states $ \left \{ h_{i},h_{i}^{Y+},h_{i}^{Z+} \right \}_{i=1}^{M} $ through the encoder $ f_{\theta } \left ( \cdot  \right ) $. Inspired by sequence level contrastive learning \cite{2022sequence}, we define the loss of $ x_{i} $ within a batch as the sum of the contrastive losses between $ \left ( x_{i},x_{i}^{Y+} \right ) $, $ \left ( x_{i},x_{i}^{Z+} \right ) $ and $ \left ( x_{i}^{Y+},x_{i}^{Z+} \right ) $ positive. The training loss function of the SIFTER-SimCSE method is as follows:

\begin{equation}
\begin{gathered}
\ell_i=-
\lambda_{X-Y} \log \frac{e^{\operatorname{sim}\left(h_i, h_i^{Y+}\right) / \tau}}
{\sum_{j=1}^N e^{\operatorname{sim}\left(h_j, h_j^{\mathrm{Y}}\right) / \tau}}-
\lambda_{X-Z} \log \frac{e^{\operatorname{sim}\left(h_i, h_i^{\mathrm{Z}+}\right) / \tau}}
{\sum_{j=1}^N e^{\operatorname{sim}\left(h_j, h_j^{\mathrm{Z}}\right) / \tau}} \\
- \lambda_{Y-Z} \log \frac{e^{\operatorname{sim}\left(h_i^{\mathrm{Y}}, h_i^{\mathrm{Z}}\right) / \tau}}
{\sum_{j=1}^N e^{\operatorname{sim}\left(h_j^{\mathrm{Y}}, h_j^{\mathrm{Z}}\right) / \tau}} \label{eq2}
\end{gathered}
\end{equation}

$\lambda _{X-Y}$, $\lambda _{X-Z}$ and $\lambda _{Y-Z}$ are weight hyperparameters. Experimental results have shown that using multiple similarity losses works better than using a single loss, which is also more efficient for training. Finally, we sum all N in-batch classification losses to obtain the final contrastive loss $\mathcal{L}_{\text {contrastive }}$.

\section{Modification of the LSTM Gate Mechanism}\label{sec4}

In this section, we will first introduce the classic Long Short-Term Memory (LSTM) model. Then, we will explore how to apply the SIFTER Strategy to the LSTM model in order to adaptively modify it for sentiment analysis tasks. This modification will allow emotional vocabulary to be directly input into the LSTM memory cells, thereby improving the model's sentence embedding performance in sentiment analysis tasks.

\subsection{LSTM Gate Mechanism}\label{sec4.1}

In the LSTM model, at the t-th time step, the data input to the LSTM unit are $x_{t}$, and its previous hidden state $h_{t-1}$ and memory cell $c_{t-1}$. The output of the LSTM unit is the memory cell $c_{t}$ and the hidden state $h_{t}$. The LSTM unit contains three gate structures, namely, a forget gate, an input gate and an output gate. The forget gate, input gate and output gate are defined as $f_{t}$, $i_{t}$ and $o_{t}$, respectively. All three gate vectors are compressed between zero and one. Intuitively, the forget gate controls how much past information should be forgotten, the input gate determines how much new information is loaded into the long-term memory cell, and the output gate allows the LSTM memory cell to select the appropriate memory for outputting. The LSTM transition equations can be written as follows:

\begin{equation}
\begin{gathered}
\mathrm{f}_t=\delta\left(W_f \cdot x_t+U_f \cdot h_{t-1}+b_t\right) \\
\mathrm{i}_t=\delta\left(W_i \cdot x_t+U_i \cdot h_{t-1}+b_t\right) \\
\mathrm{o}_t=\delta\left(W_o \cdot x_t+U_o \cdot h_{t-1}+b_t\right) \\
c_t^{+}=\tanh \left(W_c \cdot x_t+U_c \cdot h_{t-1}+b_t\right) \\
c_t=\mathrm{f}_t \odot c_{t-1}+\mathrm{i}_t \odot c_t^{+} \\
h_t=\mathrm{o}_t \odot \tanh \left(c_t\right) \label{eq3}
\end{gathered}
\end{equation}

Here, various W and U matrices are untrained parameters, and b is the bias to be learned. $\sigma$ and tanh denote the sigmoid and hyperbolic tangent functions, respectively. $\odot$ denotes the elementwise multiplication operation.

\subsection{SIFTER-LSTM}\label{sec4.2}

We describe the LSTM architecture above, and one of its limitations is that any input word must pass through an input gate to update the current memory cell. This may cause the LSTM model to perform poorly on certain downstream tasks. Unlike the semantic similarity task, in the sentiment task the polarity of a sentence is often judged by judging the words in the sentence with positive and negative meanings. If the original LSTM model is used to classify sentences, this part of the information with obvious meaning is not fully reflected. Here, we propose one extension to the basic LSTM architecture: $\mathit{SIFTER-LSTM}$. This variant allows the sentence representation to fully absorb the information contained in those important words in the sentiment analysis task and improve the resulting classification performance. 

\begin{figure}[h]
	\centering
	\includegraphics[width=0.85\linewidth]{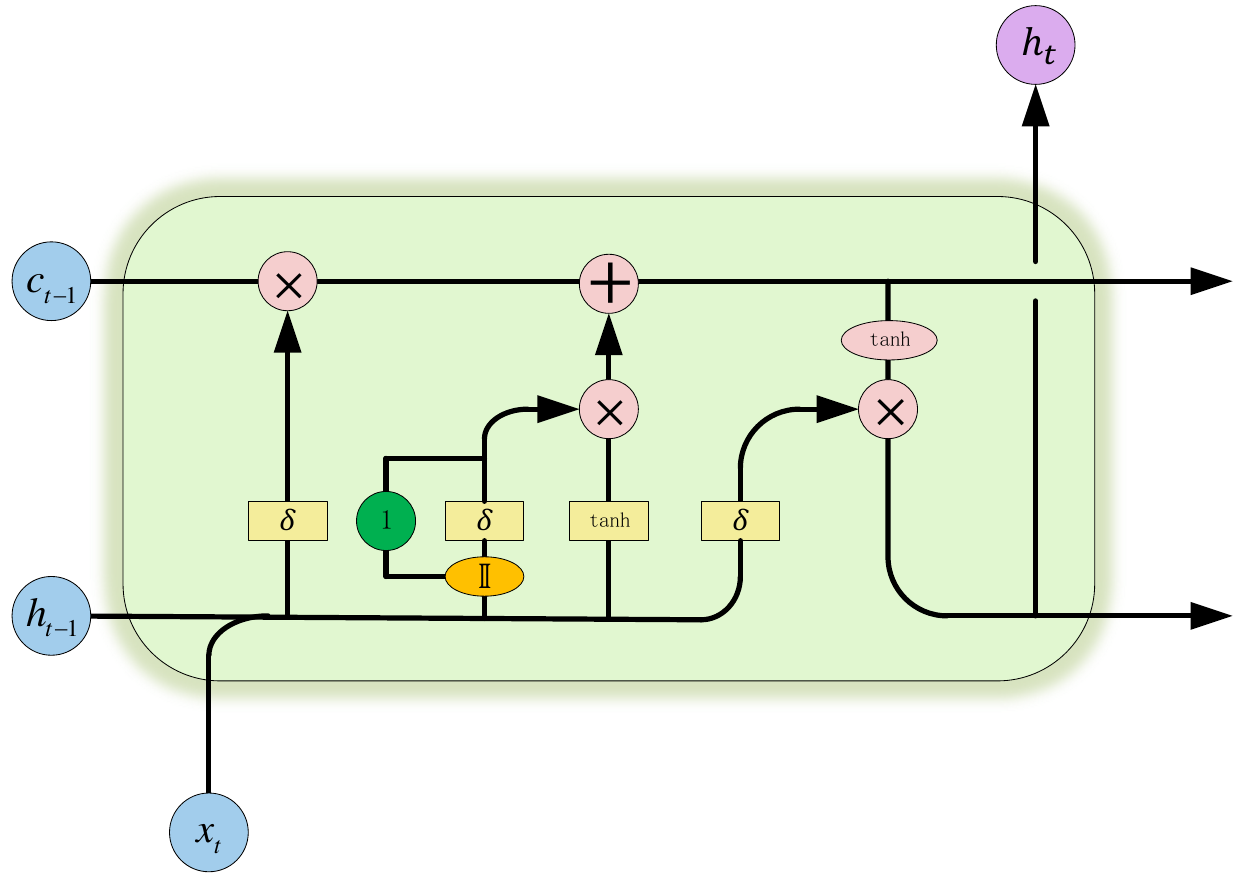}
	\caption{The architecture of SIFTER-LSTM. Compared to the standard LSTM model, our model has two additional structures, one is an indicator function and the other is a matrix of all 1s. Here, the indicator function indicates to which path, and the all-one matrix means that the input is added directly to the memory cell.}\label{<fig3>}
\end{figure}

The architecture of the SIFTER-LSTM model is shown in \cref{<fig3>}. As in the standard LSTM model, at the t-th time step, the input of the SIFTER-LSTM model is also the word vector $x_{t}$, the last hidden state $h_{t-1}$, and the last memory cell $c_{t-1}$. The difference between the standard LSTM unit and the SIFTER-LSTM unit is that the gate vector of the input gate is controlled by the indicator function. When an input word is beneficial to our downstream task, the learnable input gate parameters are short-circuited and replaced with an all-one vector of the same size. This allows the modified LSTM unit to forcibly assimilate information that is important for downstream tasks. For example, SIFTER-LSTM models can increase the memory of input words containing rich emotions while allowing them to learn normally for normal words.

In particular, when the input word $x_{t}$ belongs to the set $\mathbb{X}$ that is beneficial to downstream tasks, then the input gate becomes `short-circuited', and the input gate works normally when it does not belong. The rest of the SIFTER-LSTM is identical to the standard LSTM. The rewritten LSTM transition equations are as follows:

\begin{equation}
\begin{gathered}
\mathrm{f}_t=\delta\left(W_f \cdot x_t+U_f \cdot h_{t-1}+b_t\right)  \\
\mathrm{i}_t=\delta\left(W_i \cdot x_t+U_i \cdot h_{t-1}+b_t\right) \\
\mathrm{o}_t=\delta\left(W_o \cdot x_t+U_o \cdot h_{t-1}+b_t\right) \\
c_t^{+}=\tanh \left(W_c \cdot x_t+U_c \cdot h_{t-1}+b_t\right) \\
c_t=\mathrm{f}_t \odot c_{t-1}+\mathbb{I}\left(x_t \notin \mathbb{X}\right) \odot \mathrm{i}_t \odot c_t^{+}+\mathbb{I}\left(x_t \in \mathbb{X}\right) \odot c_t^{+} \\
h_t=o_t \odot \tanh \left(c_t\right) \label{eq4}
\end{gathered}
\end{equation}

Here, $\mathbb{I}$ represents the indicator function. It takes a value of 1 when the statement in brackets is true; otherwise the function takes a value of 0.

\subsection{SIFTER-LSTM for text classification}\label{sec4.3}

For the text classification task, given an input sequence X of length T: $\left \{ x_{1},\cdots, x_{t} \right \} $ and its true label $ y\in \left \{ 1,2,\cdots,k\right \} $, we feed the sequence into the modified model, and then we can obtain T hidden states: $\left \{ h_{1},\cdots, h_{t} \right \}$. We treat the output $h_{t}$ as the whole representation of the input. It is then passed through a linear transformation and a softmax layer to estimate the probability $\hat{y}$ for each label $Y\in \left \{ 1,2,\cdots,k \right \} $. The classifier takes the hidden state $h_{T}$ as input:

\begin{equation}
\begin{gathered}
\tilde{p}(y \mid X)=\operatorname{softmax}\left(W \cdot h_t+b\right) \\
\hat{y}=\arg \max _y \tilde{p}(y \mid X) \label{eq5}
\end{gathered}
\end{equation}

We train the model using the cross-entropy loss minimization. The loss function is defined as:

\begin{equation}
\mathcal{J}(\theta)=-\sum_{i=1}^k \mathbb{I}(\mathrm{x}==\mathrm{y}) \log \left(y_i\right)+\lambda\|\theta\|_2^2\label{eq6}
\end{equation}

where $y_{i}$ is the probability for the i-th class estimated by softmax function, k is the number of text classes, and $\lambda$ is an L2 regularization hyperparameter. The training details are further introduced in \cref{sec5.4}.

\section{Experiments}\label{sec5}

\subsection{Evaluation Setup}\label{sec5.1}

We conduct experiments on 7 STS tasks and 1 sentiment analysis task. In the former tasks, we mainly follow the ESimCSE setting and use the SentEval toolkit to evaluate sentence embeddings. For the latter task, we leverage movie reviews to evaluate the performance of the SIFTER-LSTM model.

\subsection{Semantic Textual Similarity}\label{sec5.2}

Datasets commonly used for text similarity tasks include STS 12-16 \cite{2012semeval, 2013sem, 2014semeval, 2015semeval, 2016semeval}, STS-Benchmark \cite{2017semeval}, SICK-Relatedness \cite{ 2014sick}. Since the STS 12-16 datasets do not have training and development sets, we randomly sample 1 million sentences from Wikipedia for model training and select hyperparameters and model checkpoints based on the development set of STS-B. After the model training process is complete, we compute the cosine similarity between each pair of sentences for all test sets and finally use spearman correlation to measure the correlation between a given gold label and the predicted similarity. We use the SentEval toolkit to complete the above process.

\subsection{Sentiment Classification}\label{sec5.3}

We evaluate the performance of the SIFTER-LSTM model using the Stanford Sentiment Treebank (SST) dataset \cite{2013recursive}. In this paper, we only evaluate on the binary classification task on this dataset. The binary classification data set train/dev/test is divided into 67349/872/1821. We train the improved model on the training set, select the optimal parameters on the development set, and finally evaluate the performance of the model on the test set. 

\subsection{Training Details}\label{sec5.4}

For the semantic similarity tasks, we randomly sample 1 million sentences from Wikipedia for model training, but not every sentence has a backbone, and each sentence may have grammatical errors. Therefore, our crude approach is to remove sentences that are less than three words long from these data and remove sentences with capitalized last words and their next sentence. We start with a checkpoint of pretrained models provided by the HuggingFace's transformers package, which includes the BERT \cite{Bert} and RoBERTa \cite{2019roberta} base models and two corresponding larger versions. 

For the SIFTER-SimCSE model, we take a [CLS] vector with multilayer perceptron (MLP) layers on top of it as the sentence representations. Similar to SimCSE, we also only add the MLP layer at training time, discard it at test time and use the [CLS] vector for evaluation purposes. The model hyperparameters essentially follows the setup of the ESimCSE paper. When training the model, we train for one epoch using the AdamW optimizer on two A100 GPUs with learning rate = 1e-5 and batch size = 64 for all models. The three $\lambda$ parameters in \cref{eq2} are all 1, which means that our model uses three contrastive losses to jointly the optimize the pretrained language encoder. The temperature $\tau$ is set to 0.05, and the dropout rate is p = 0.15 for all the experiments. Finally, we evaluate the model every 125 training steps on the development set of STS-B and reserve the best checkpoints for the final evaluation on the test set.

For the sentiment classification task, in the experiments we found that the words that really affect the classification performance of the model belong to the words with positive and negative meanings. So we want to allow these words to be directly fed into the memory cell without the input gate. 

In the experimental setup, we select the optimal model parameters based on the test set of the SST-2 sentiment dataset and use a single NVIDIA P100 GPU for this experiment. We use the same WordPiece embeddings \cite{2016google} as those of the BERT model to initialize the word vectors. The word representations can be updated during training. The dimensionality of the input word vector is 768, and the SIFTER-LSTM hidden layer has 384 dimensions. The word vector layer and the LSTM layer are dropped out with a probability of 0.2. Our model is trained using the Adam optimizer with a learning rate of 1e-5 and a min-batch size of 32. The model parameters use an L2 penalty with a factor of 1e-7. During training, we conduct a validation the development set every 50 training steps, saving the best model checkpoint. For a fair comparison, we use the same parameter settings to reproduce the effects of the original LSTM model.

\subsection{Results}\label{sec5.5}

$\mathbf{Semantic}$ $\mathbf{Textual}$ $\mathbf{Similarity}$ $\mathbf{(STS)}$. We compare the SIFTER-SimCSE model with many strong unsupervised baselines including SimCSE \cite{Simcse} and ESimCSE \cite{ESimcse}. We will next show the evaluation performance of our model on 7 STS datasets. As shown in \cref{tab3}, our method outperforms the previously developed methods under four different parameter scale settings. SIFTER-SimCSE-BERT$_{base}$ can significantly outperform SimCSE-BERT$_{base}$ and raise the averaged Spearman's correlation from 76.25\% to 78.20\%. The effect of switching to the RoBERTa encoder is even greater, with our method reaching 78.54\% using the RoBERTa$_{base}$ model.

\begin{table*}[t]
	\begin{center}
		\centering
		\small
		
		\caption{
			Performance of different sentence representation models on the STS dataset (Spearman's correlation).
			$\clubsuit$: results from SimCSE\cite{Simcse};
			$\spadesuit$: results from ESimCSE\cite{ESimcse};
			Other results are reproduced by us.
		}\label{tab3}
		
		\scalebox{0.75}{
		\begin{tabular}{lcccccccc}
			\toprule
			{Model} & {STS12} & {STS13} & {STS14} & {STS15} & {STS16} & {STS-B} & {SICK-R} & {Avg.} \\
			\midrule
			\midrule
			SimCSE-BERT$_{base}$$\clubsuit$ & 68.4 & 82.41 & 74.38 & 80.91 & 78.56 & 76.85 & 72.23 & 76.25 \\
			ESimCSE-BERT$_{base}$ $\spadesuit$  & \textbf{73.4} & 83.27 & \textbf{77.25} & 82.66 & 78.81 & \textbf{80.17} & 72.3 & \textbf{78.27} \\
			*SIFTER-SimCSE-BERT$_{base}$ & 72.2 & \textbf{84.46} & 76.05 & \textbf{82.97} & \textbf{79.09} & 79.65 & \textbf{73.01} & 78.2 \\
			\midrule
			SimCSE-BERT$_{large}$$\spadesuit$ & 70.88 & 84.16 & 76.43 & 84.5 & 79.76 & 79.26 & 73.88 & 78.41 \\
			ESimCSE-BERT$_{large}$ $\spadesuit$ & \textbf{73.21} & 85.37 & 77.73 & 84.3 & 78.92 & 80.73 & \textbf{74.89} & 79.31 \\
			*SIFTER-SimCSE-BERT$_{large}$  & 72.73 & \textbf{86.14} & \textbf{78.06} & \textbf{84.94} & \textbf{80.5} & \textbf{80.74} & 73.88 & \textbf{79.57} \\
			\midrule
			SimCSE-RoBERTa$_{base}$ $\clubsuit$& 70.16 & 81.77 & 73.24 & 81.36 & 80.65 & 80.22 & 68.56 & 76.57 \\
			ESimCSE-RoBERTa$_{base}$ $\spadesuit$ & 69.9 & 82.5 & 74.68 & \textbf{83.19} & 80.3 & 80.99 & 70.54 & 77.44 \\
			*SIFTER-SimCSE-RoBERTa$_{base}$  & \textbf{71.61} & \textbf{83.63} & \textbf{76.56} & 82.78 & \textbf{81.94} & \textbf{82.61} & \textbf{70.64} & \textbf{78.54} \\
			\midrule
			
			SimCSE-RoBERT$_{large}$ $\clubsuit$ & 72.86 & 83.99 & 75.62 & 84.77 & 81.8 & 81.98 & 71.26 & 78.9 \\
			ESimCSE-RoBERTa$_{large}$ $\spadesuit$ & 73.2 & 84.93 & 76.88 & \textbf{84.86} & 81.21 & 82.79 & \textbf{72.27} & 79.45 \\
			*SIFTER-SimCSE-RoBERTa$_{large}$ & \textbf{73.71} & \textbf{85.5} & \textbf{78.46} & 84.46 & \textbf{81.98} & \textbf{83.21} & 71.58 & \textbf{79.84} \\
			
			\bottomrule
		\end{tabular}}
	\end{center}
	
	\label{tab:replicated_results}
\end{table*}

$\mathbf{Sentiment}$ $\mathbf{Analysis}$ $\mathbf{(SA)}$.The results of our method are summarized in \cref{tab4}. We also reproduce the experimental results of the standard LSTM model with the same parameters. The SIFTER-LSTM achieves 85.37\% accuracy on the binary sentiment classification task, outperforming the 84.90\% accuracy of the standard LSTM model. This shows the effectiveness of the SIFTER idea in terms of model improvement guidance.

\begin{table}[htbp]
	\centering
	\caption{Classification accuracies achieved on the SST dataset. $\diamondsuit$: results from Tree-LSTM. In our experiments, we fix the random seed and report the mean accuracy over 5 runs, where the numbers in parentheses represent the standard deviations. SST-2: positive/negative sentiment classification.}\label{tab4}
	\begin{tabular}{p{20.69em}p{9.94em}}
		\toprule
		Method & SST-2 \\
		\midrule
		LSTM$^{\diamondsuit }$ & 84.9(0.6) \\
		LSTM(our implementation) & 84.97(0.38) \\
		SIFTER-LSTM & 85.37(0.31) \\
		\bottomrule
	\end{tabular}%
	
\end{table}%

\section{Ablation Studies}\label{sec6}

In this section, for the semantic text similarity task, we investigate the effects of different dropout rates and different $\lambda$ parameters in \cref{eq2}. All reported results are based on the STS-B development set. For sentiment classification, we also explore the effects of word embeddings on the LSTM variants. In the following ablation experiments, we modify one hyperparameter at a time, leaving the other settings unchanged. 

\subsection{The Impact of Coefficient $\lambda$}\label{sec6.1}

In \cref{eq2}, we use different $\lambda$ $\left ( \mathrm{i.e}.,\lambda_{X-Y} \right )$ coefficient parameters to weight the loss between the three positive sample pairs. Note that SIFTER-SimCSE $\left ( \lambda_{\ast-\ast} \right )$ means that $ \lambda_{\ast-\ast}  = 1$ and all the other $\lambda s$ are equal to zero in the equation. 

For a fair comparison, we take the same experimental setup as ESimCSE when $\lambda $ is taken to be 1, that is, learning rate is 3e-5 and batch size is 64. We can see that SIFTER-SimCSE $\left ( \lambda_{X-Y} \right )$, $\left ( \lambda_{X-Z} \right )$ and $\left ( \lambda_{Y-Z} \right )$ all significantly outperform SimCSE, which demonstrates the effectiveness of our proposed data augmentation method. In particular, the best results are obtained when we use three sentence pairs for contrastive learning. This phenomenon indicates that our method indeed makes the pretrained model pay more attention to the meaningful parts of the given sentence for the STS task.

\begin{table}[htbp]
	\centering
	\caption{SIFTER-SimCSE $\left ( \lambda_{X-Y} \right )$, SIFTER-SimCSE $\left ( \lambda_{X-Z} \right )$ and SIFTER-SimCSE $\left ( \lambda_{Y-Z} \right )$ respectively stand for contrastive learning between the original sentence and the sentence obtained after adding the main information, the original sentence and the sentence obtained after deleting the useless information, and the sentence obtained after adding the main information and the sentence obtained after deleting the useless information.} \label{tab5}
	
	\begin{tabular}{p{27.25em}c}
		\toprule
		Model & \multicolumn{1}{p{2.875em}}{STS-B} \\
		\midrule
		unsup-SimCSE-BERT$_{base}$ & 82.45 \\
		SIFTER-SimCSE-BERT$_{base}\left ( \lambda_{X-Y}   \right )$ & 83.35 \\
		SIFTER-SimCSE-BERT$_{base}\left ( \lambda_{X-Z}   \right )$ & 83.92 \\
		SIFTER-SimCSE-BERT$_{base}\left ( \lambda_{Y-Z}   \right )$ & 83.72 \\
		SIFTER-SimCSE-BERT$_{base}\left ( \lambda_{X-Y}+\lambda_{X-Z}+\lambda_{Y-Z}  \right )$ & 84.39 \\
		\bottomrule
	\end{tabular}%
\end{table}%

\subsection{The Impact of Dropout Rate}\label{sec6.2}

In SimCSE, the authors viewed embeddings of the same sentence with different dropout masks as a minimal form of data augmentation. Therefore, using different dropout rates does affect the performance of the encoder. We experiment with three different dropout rates and the results are shown in \cref{tab6}. Unlike SimCSE, the SIFTER-SimCSE-BERT$_{base}$ model achieves the best performance on the STS-B development set when the dropout is 0.15 and $\lambda_{X-Y}=\lambda_{X-Z}=\lambda_{Y-Z}=$ 1 in \cref{eq2}. In terms of dropout, our experimental results show that the proposed model outperforms SimCSE.

\begin{table}[htbp]
	\centering
	\caption{STS-B development set results of the SIFTER-SimCSE-BERT$_{base}$ model under different dropout probabilities $\mathit{p}$.}\label{tab6}%
	\begin{tabular}{p{15.5em}ccc}
		\toprule
		\textit{p} & 0.1 & 0.15 & 0.2 \\
		STS-B & 83.94 & 84.39 & 82.81 \\
		\bottomrule
	\end{tabular}%
	
\end{table}%

\subsection{Stability of Contrastive Learning}\label{sec6.3}

To demonstrate the effectiveness of our method, we take the experimental results obtained with 10 random seeds and compare them with those of the unsupervised SimCSE method. For the results of SimCSE, we take the experimental results presented in the paper on PromptBERT \cite{2022promptbert}. From \cref{tab7}, we can see that our method is more stable than SimCSE. Even the worst model performs on par with the best SimCSE model. In the SIFTER-SimCSE model, the difference between the best and worst results is 1.76\%.

\begin{table}[htbp]
	\centering
	\caption{Experimental results obtained by using unsupervised contrastive learning with 10 random seeds on the STS-B test set.$\heartsuit$:results from PromptBERT.}\label{tab7}
	\begin{tabular}{p{20.75em}ccc}
		\toprule
		Model & \multicolumn{1}{p{2.19em}}{Max} & \multicolumn{1}{p{2.19em}}{Min} & \multicolumn{1}{p{2.19em}}{Mean} \\
		\midrule
		unsup-SimCSE-BERT$_{base}$ $\heartsuit$ & 76.64 & 73.5 & 75.42 \\
		SIFTER-SimCSE-BERT$_{base}$ & 78.2 & 76.44 & 77.09 \\
		\bottomrule
	\end{tabular}%
	\label{tab:addlabel}%
\end{table}%

\section{Conclusion}\label{sec7}

In this study, we propose the idea of enhancing the information in relevant parts of downstream tasks and reducing the irrelevant elements of tasks in different domains (SIFTER). We show how to incorporate semantic similarity task features with the SimCSE approach and modify the LSTM model via guidance provided by features from the sentiment analysis task. Empirical improvements achieved on two different domain tasks demonstrate the effectiveness and transferability of the SIFTER strategy. We believe that it is a very good practice to improve specific methods through domain-specific features. In the future, we will further explore its potential in more tasks, such as text retrieval.

\bibliographystyle{sn-mathphys}
\bibliography{reference}

\end{document}